\documentclass[11pt]{article}

\usepackage[utf8]{inputenc}
\usepackage[T1]{fontenc}
\usepackage{graphicx}
\usepackage{amsmath,amssymb,amsfonts}
\usepackage{hyperref}
\usepackage{url}
\usepackage{booktabs}
\usepackage{multirow}
\usepackage{array}
\usepackage{longtable}
\usepackage{rotating}
\usepackage{pdflscape}
\usepackage[title]{appendix}

\usepackage[margin=1in]{geometry}

\hypersetup{
    colorlinks=true,
    linkcolor=blue,
    filecolor=magenta,
    urlcolor=[RGB]{0,0,128},
    citecolor=blue,
}

\title{The Massive Legal Embedding Benchmark (MLEB)}

\author{
    Umar Butler\thanks{Corresponding author: \texttt{research@isaacus.com}}, Abdur-Rahman Butler, Adrian Lucas Malec \\
    Isaacus
}

\date{21 October 2025}

\begin{document}

\maketitle

\begin{abstract}
We present the \textbf{Massive Legal Embedding Benchmark (MLEB)}\footnote{\url{https://isaacus.com/mleb}}, the largest, most diverse, and most comprehensive open-source benchmark for legal information retrieval to date. MLEB consists of ten expert-annotated datasets spanning multiple jurisdictions (the US, UK, EU, Australia, Ireland, and Singapore), document types (cases, legislation, regulatory guidance, contracts, and literature), and task types (search, zero-shot classification, and question answering). Seven of the datasets in MLEB were newly constructed in order to fill domain and jurisdictional gaps in the open-source legal information retrieval landscape. We document our methodology in building MLEB and creating the new constituent datasets, and release our code, results, and data openly to assist with reproducible evaluations.
\end{abstract}

\section{Introduction}
In the context of information retrieval, embedding models convert documents and queries into sets of numbers known as `embeddings' that can be compared with each other to identify relevant search results. Embeddings power the `retrieval' component of retrieval-augmented generation (RAG) applications and are widely used across the legal tech industry. In legal RAG applications, low-quality embeddings lead to low-quality search results, which in turn lead to low-quality responses and increased hallucinations~\cite{acl2025rag}.

Despite their importance, limited attention has been paid to ensuring that embedding models are genuinely fit for legal information retrieval. Previous attempts at building an industry-standard legal information retrieval benchmark have been limited in quality, size, and diversity. LegalBench-RAG~\cite{legalbenchrag2024} focuses on a small set of contract- and US-centric datasets, while the legal domain subset of the Massive Multilingual Text Embedding Benchmark (MTEB-Legal)~\cite{mteb2025} exhibits labeling issues and narrow topical coverage. Consequently, existing benchmark performance can fail to predict real-world effectiveness at legal retrieval tasks. MLEB attempts to address those limitations by being larger, more diverse, and of higher quality than previous legal information retrieval benchmarks.

\section{Related work}

\subsection{LegalBench-RAG}
LegalBench-RAG evaluates legal information retrieval performance against four pre-existing evaluation sets: ContractNLI~\cite{contractnli}, Contract Understanding Atticus Dataset (CUAD)~\cite{cuad2021}, M\&A Understanding Dataset (MAUD)~\cite{maud2023}, and Privacy QA~\cite{privacyqa}. All datasets center around contracts, largely sourced from the US.

In practice, legal professionals and users seeking legal advice or knowledge tend to search for and be interested in a much broader range of document types than just contracts, including legislation, regulations, cases, and general legal literature.

LegalBench-RAG's narrow focus on contracts thus limits its usefulness for the evaluation of legal embedding models that generalize well to other legal domains and multiple jurisdictions.

\subsection{MTEB-Legal}
MTEB-Legal consists of eight datasets: AILA Casedocs~\cite{aila2019}, AILA Statutes~\cite{aila2019}, GerDaLIR Small~\cite{gerdalir}, LeCaRDv2~\cite{lecardv2}, Consumer Contracts QA~\cite{kolt2022predicting}, Legal Summarization~\cite{legalsummarization}, Corporate Lobbying~\cite{corporatelobbying}, and LegalQuAD~\cite{legalquad}.

Upon manual inspection of two of the English-language datasets in MTEB-Legal, AILA Casedocs\footnote{\url{https://huggingface.co/datasets/mteb/AILA_casedocs}} and AILA Statutes\footnote{\url{https://huggingface.co/datasets/mteb/AILA_statutes}}, we found they contained many query-passage pairs that were totally irrelevant to each other. According to the authors, the datasets had been created using an `automated methodology' that paired `facts stated in certain [Indian] Supreme Court cases' with cases and statutes that had been `cited by the lawyers arguing those cases'. This construction method was employed because `actually involving legal experts (e.g., to find relevant prior cases / statutes) would have required a significant amount of financial resources and time'~\cite{aila2019}.

In addition to mislabeling, we found that MTEB-Legal lacked diversity in the areas that matter most to legal practitioners and seekers of legal knowledge.

Specifically, of the remaining English-language datasets after exclusion of AILA Casedocs and AILA Statutes, two deal with consumer terms of service (Consumer Contracts QA and Legal Summarization), leaving only one (Corporate Lobbying) that deals with legislation, and none dealing with case law. All such datasets are largely representative of American law.

Regarding the non-English-language datasets in the legal split of MTEB, we argue that, in many cases, the legal systems of different cultures may fundamentally differ in ways that make cross-jurisdictional comparisons (e.g., between the common law system used by Anglosphere countries and Sharia law) of the effectiveness of legal embeddings inappropriate.

Furthermore, given that the legal split contains two German datasets, one Chinese dataset, and no other non-English datasets, and that those datasets are concentrated on three select legal tasks, we argue that the inclusion of non-English datasets largely introduces bias and noise in ways that are unlikely to be conducive to real-world performance on most English-language legal information retrieval tasks.

\section{The Massive Legal Embedding Benchmark (MLEB)}
Learning from the limitations of existing legal embedding evaluation sets, MLEB has been designed with four key objectives in mind, namely to:

\begin{enumerate}
    \item be of high quality, both in terms of provenance and labeling;
    \item consist of text processing tasks that have genuine real-world utility to legal tech professionals;
    \item be meaningfully challenging in ways likely to require significant legal knowledge and strong legal reasoning skills; and
    \item represent a broad variety of jurisdictions, legal areas, and types of legal texts.
\end{enumerate}

To that end, MLEB contains ten different evaluation sets spanning a range of difficulties (including tasks requiring legal reasoning as well as tasks requiring lexical analysis), problem types (specifically, retrieval, zero-shot classification, and question answering), jurisdictions (the US, UK, EU, Australia, Ireland, and Singapore), and document types (decisions, legislation, regulatory guidance, contracts, and literature).

Of the ten datasets in MLEB, seven are entirely new, constructed either by having subject matter experts hand-label data or by adapting existing expert-labeled data.

Below, we present an overview of all the datasets included in MLEB alongside all the various features that make them unique.

\begin{table}[h!]
\centering
\small
\begin{tabular}{>{\raggedright}p{2.5cm}>{\raggedright}p{1.1cm}>{\raggedright}p{1.8cm}>{\raggedright\arraybackslash}p{2.1cm}p{6.5cm}}
\toprule
\textbf{Name} & \textbf{Queries} & \textbf{Domain} & \textbf{Jurisdiction} & \textbf{Description} \\
\midrule
Bar Exam QA & 117 & Judicial & US & US bar exam questions paired with relevant caselaw.~\cite{Zheng_2025} \\
\addlinespace
SCALR & 120 & Judicial & US & Questions presented to the US Supreme Court paired with descriptions of the Court's final holdings.~\cite{scalr} \\
\addlinespace
Singaporean Judicial Keywords & 500 & Judicial & Singapore & Judicial catchwords paired with Singaporean court judgments. \\
\addlinespace
GDPR Holdings Retrieval & 500 & Judicial & EU & GDPR case fact patterns paired with descriptions of court holdings. \\
\addlinespace
Australian Tax Guidance Retrieval & 112 & Regulatory & Australia & Australian tax law questions paired with relevant Australian Government tax guidance and policies. \\
\addlinespace
Irish Legislative Summaries & 500 & Regulatory & Ireland & Long titles paired with Irish acts. \\
\addlinespace
UK Legislative Long Titles & 78 & Regulatory & UK & Long titles paired with UK acts. \\
\addlinespace
Contractual Clause Retrieval & 90 & Contractual & Multinational & NLI-style descriptions of types of contractual clauses paired with examples of those clauses. \\
\addlinespace
License TL;DR Retrieval & 65 & Contractual & Multinational & Summaries of software licenses paired with their full texts. \\
\addlinespace
Consumer Contracts QA & 198 & Contractual & Multinational & Questions about online terms of service paired with relevant clauses.~\cite{kolt2022predicting} \\
\bottomrule
\end{tabular}
\caption{An overview of the ten datasets making up MLEB, spanning six jurisdictions (the US, UK, EU, Australia, Ireland, and Singapore) and three domain types (Judicial, Contractual, Regulatory). Full URLs are provided in the Data Availability section (page~\pageref{sec:data_availability}).}
\label{tab:mleb_datasets}
\end{table}

\subsection{Bar Exam QA}
Bar Exam QA~\cite{barexamqa} is a pre-existing evaluation set created by Stanford RegLab that tests the ability of information retrieval models to identify cases and legal literature (specifically, passages from textbooks) relevant to US state bar exam questions. Bar Exam QA was selected to be a challenging retrieval task requiring lawyer-like analytical skills and extensive knowledge of American law for an information retrieval model to achieve high performance.

\subsection{SCALR}
SCALR~\cite{scalr} is a pre-existing evaluation set created by Faiz Surani and Varun Iyer that pairs questions presented to the US Supreme Court with descriptions of the Court's final holdings. Uniquely, SCALR challenges models to predict, or at least be aware of, important US Supreme Court holdings. Half of SCALR was reserved for validation, leaving the other half for inclusion in the MLEB test set, with no queries overlapping between sets.

\subsection{Singaporean Judicial Keywords}
Singaporean Judicial Keywords is a new evaluation set consisting of 500 catchword-judgment pairs sourced from the Singaporean Judiciary.

Singaporean Judicial Keywords was constructed by collecting all publicly available Singaporean court judgments, converting them into plain text with Inscriptis~\cite{inscriptis}, cleaning them, and removing near duplicates with the simhash algorithm~\cite{simhash}, and then using multiple complex regex patterns to extract catchwords from them before removing those catchwords and everything preceding them from judgments (in order to force models to focus on representing the core semantics of judgments' texts rather than their metadata-rich cover sheets).

Uniquely, the keywords in this dataset are real-world annotations created by subject matter experts, namely, Singaporean law reporters, as opposed to being constructed ex post facto by third parties.

Additionally, unlike standard keyword queries, judicial catchwords are meant to capture the most essential and relevant concepts and principles to a case, even where those elements may never be explicitly referenced by it.

Such features make this dataset especially useful for the robust evaluation of the legal conceptual understanding and overall legal knowledge of information retrieval models.

\subsection{GDPR Holdings Retrieval}
GDPR Holdings Retrieval is a new evaluation set consisting of 500 fact patterns paired with holdings in European regulatory and court decisions. This dataset was constructed by collecting all GDPRHub~\cite{gdprhub} articles, using regex to separate their facts and holdings sections and then converting those sections into plain text with Inscriptis. This dataset is intended to stress test the ability of an information retrieval model to retrieve relevant judicial and regulatory decisions given an arbitrary fact pattern.

\subsection{Australian Tax Guidance Retrieval}
Australian Tax Guidance Retrieval is a new evaluation set consisting of 112 real-life tax questions posed by Australian taxpayers paired with 105 relevant Australian Government guidance and policy documents.

Questions in this dataset were sourced from the Australian Taxation Office (ATO)'s community forum where Australian taxpayers can ask accountants and ATO officials their tax questions. We found that, in many cases, users' questions can be answered by reference to Australian Government guidance materials that, for whatever reason, taxpayers were unable to locate themselves.

We thus constructed this dataset by:
\begin{enumerate}
    \item for each of the 14 sub-topics of the ATO Community forum that did not come under the parent topics `Online Services' and `Tax Professionals' (which were found to consist almost exclusively of practical questions around the use of ATO services rather than substantive tax law queries), selecting 8 questions that:
    \begin{enumerate}
        \item had at least one answer with at least one hyperlink (with, where there were multiple competing answers, the answer selected by the user as the best answer being used otherwise using the answers of ATO employees over those of tax professionals),
        \item were about a substantive tax law problem and were not merely practical questions about, for example, the use of ATO services or how to file tax returns;
    \end{enumerate}
    \item for each sampled question, visiting the hyperlink in the selected answer that appeared to be the most relevant to the question and then copying as much text from the hyperlink as appeared relevant to the question, ranging from a single paragraph to the entire document;
    \item using a purpose-built Chrome extension to extract questions and relevant passages directly to Markdown to preserve the semantics of added markup; and
    \item lightly cleaning queries and passages by replacing consecutive sequences of at least two newlines with two consecutive newlines and removing leading and trailing whitespace.
\end{enumerate}

The queries in this dataset are valuable and challenging because users have gone to the effort of asking them on a forum, indicating that traditional search engines failed to surface the answers they were looking for. The government materials are, in turn, also valuable because practicing subject matter experts, namely, accountants and ATO officials, have confirmed them to be relevant.

\subsection{Irish Legislative Summaries}
Irish Legislative Summaries is an evaluation set consisting of 500 Irish laws and their long titles, succinctly summarizing the subject matter, scope, and purpose of legislation. Similar to Singaporean Judicial Keywords, this dataset was constructed by collecting all publicly available Irish laws, converting them into plain text with Inscriptis, cleaning them, and removing near duplicates with the simhash algorithm, and then using regex to extract their long titles before removing those long titles and everything preceding them from legislation (in order to force models to focus on representing the core semantics of acts' texts rather than their metadata-rich front matter). This dataset is meant to stress test the ability of an information retrieval model to retrieve relevant statutes to short queries describing them.

\subsection{UK Legislative Long Titles}
UK Legislative Long Titles is an evaluation set consisting of 78 UK laws and their long titles. This dataset was constructed from all publicly available UK laws in the same way as Irish Legislative Summaries.

\subsection{Contractual Clause Retrieval}
Contractual Clause Retrieval is an evaluation set consisting of 45 unique types of contractual clauses paired with 2 highly representative examples of each, resulting in 90 pairings.

This dataset was constructed by:
\begin{enumerate}
    \item coming up with 45 contractual clause types and NLI-style statements (see Appendix~\ref{app:clause_types}), which were intended to be as representative of the diverse types of contractual clauses common in commercial transactions while also being specific and distinct enough to avoid capturing substantively overlapping clause types. These included set-off clauses, authority to sign clauses, power of attorney, termination for cause clauses, good faith clauses, payment currency clauses, and many others;
    \item for each clause type, sourcing two highly representative examples from online clause reference databases, stripping out irrelevant information where that might cause overlap with another clause type.
\end{enumerate}

This dataset is intended to stress test the ability of information retrieval, zero-shot classification, and NLI models to identify a broad range of common types of contractual clauses based solely on their definition, without prior examples.

\subsection{License TL;DR Retrieval}
License TL;DR Retrieval is an evaluation set consisting of 65 summary-license pairs sourced from tl;drLegal. This dataset was constructed by collecting all licenses publicly available on tl;drLegal and pairing their human-created summaries with their full texts. This dataset is intended to stress test the ability of an information retrieval model to match relevant open-source licenses with summaries of their terms.

\subsection{Consumer Contracts QA}
Consumer Contracts QA~\cite{kolt2022predicting} is a pre-existing evaluation set created by Noam Kolt that tests the ability of information retrieval models to retrieve relevant contractual clauses to questions about consumer contracts. The MLEB version of this dataset was derived by splitting the MTEB version in half so that some examples could be reserved for validation, with no queries overlapping between sets.

\section{Results}

\subsection{Performance}
As of 21 October 2025, Isaacus' Kanon 2 Embedder legal embedding model ranks first on MLEB out of 20 other models, with an NDCG@10 score of 86.03, followed by Voyage 3 Large at 85.71 and Voyage 3.5 at 84.07.

The full results of the benchmark are presented below. All scores are NDCG@10 scores. We report both the task average (i.e., by evaluation set) and the domain average.

\begin{landscape}
\begin{table}[h!]
\centering
\small
\begin{tabular}{clcccccc}
\toprule
\textbf{Rank} & \textbf{Model} & \textbf{Task avg} & \textbf{Domain avg} & \textbf{Judicial} & \textbf{Contractual} & \textbf{Regulatory} \\
\midrule
1 & Kanon 2 Embedder & \textbf{86.03} & \textbf{86.37} & \textbf{82.96} & 84.67 & \textbf{91.48} \\
2 & Voyage 3 Large & 85.71 & 86.08 & 82.33 & \textbf{86.05} & 89.87 \\
3 & Voyage 3.5 & 84.07 & 84.62 & 79.13 & 85.92 & 88.83 \\
4 & Qwen3 Embedding 8B & 82.96 & 83.45 & 78.49 & 82.88 & 88.99 \\
5 & Voyage 3.5 Lite & 82.41 & 82.96 & 77.46 & 83.21 & 88.19 \\
6 & Qwen3 Embedding 4B & 81.96 & 82.61 & 76.14 & 84.19 & 87.49 \\
7 & Gemini Embedding & 80.90 & 81.50 & 75.49 & 77.88 & 91.13 \\
8 & Voyage Law 2 & 79.63 & 79.75 & 78.55 & 81.65 & 79.07 \\
9 & Text Embedding 3 Large & 78.91 & 79.31 & 75.24 & 79.70 & 83.00 \\
10 & Jina Embeddings v4 & 78.62 & 79.16 & 73.78 & 76.44 & 87.26 \\
11 & Qwen3 Embedding 0.6B & 77.13 & 77.66 & 72.37 & 76.61 & 84.01 \\
12 & EmbeddingGemma & 75.16 & 76.08 & 66.85 & 76.07 & 85.32 \\
13 & Snowflake Arctic Embed L v2.0 & 74.08 & 74.70 & 68.50 & 76.41 & 79.17 \\
14 & Snowflake Arctic Embed M v2.0 & 73.94 & 74.60 & 67.99 & 73.36 & 82.46 \\
15 & Text Embedding 3 Small & 73.88 & 74.65 & 67.00 & 74.24 & 82.71 \\
16 & Text Embedding Ada 002 & 72.59 & 73.62 & 63.29 & 73.59 & 83.99 \\
17 & Granite Embedding English R2 & 72.45 & 73.12 & 66.48 & 69.14 & 83.73 \\
18 & BGE M3 & 69.44 & 70.60 & 59.00 & 73.04 & 79.75 \\
19 & Mxbai Embed Large v1 & 69.33 & 70.50 & 58.81 & 72.10 & 80.60 \\
20 & E5 Large Instruct & 68.11 & 69.00 & 60.04 & 65.62 & 81.36 \\
21 & Granite Embedding Small English R2 & 68.06 & 68.88 & 60.67 & 64.49 & 81.48 \\
\bottomrule
\end{tabular}
\caption{Performance of 21 embedding models on MLEB. All scores are NDCG@10 values. Task average represents the mean across all ten evaluation sets; domain average represents the mean across domain types (weighted by number of datasets per domain). Performance is also shown broken down by the three domain types: Judicial, Contractual, and Regulatory.}
\label{tab:results}
\end{table}
\end{landscape}

\begin{figure}[h!]
\centering
\includegraphics[width=0.9\textwidth]{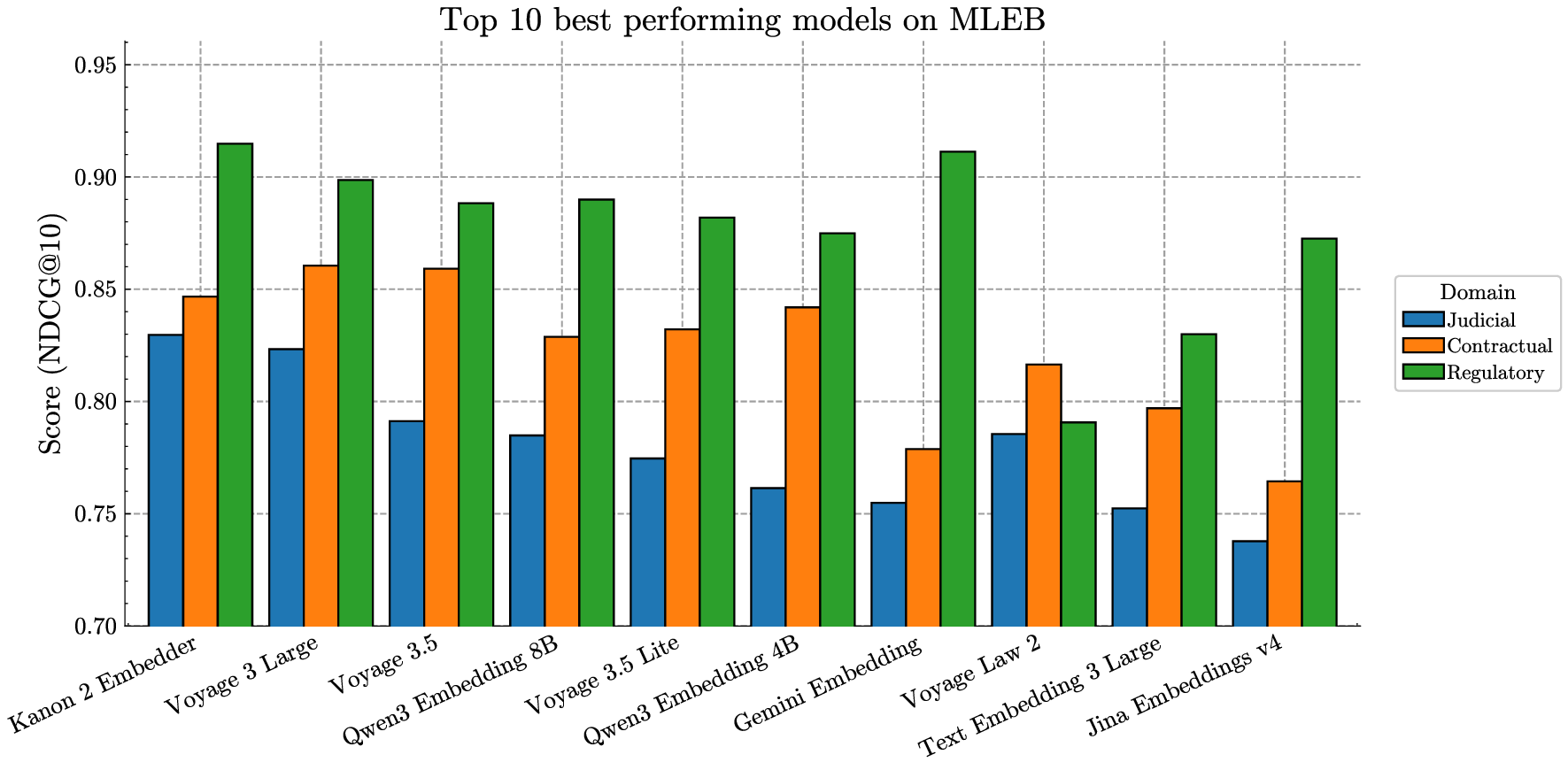}
\caption{Performance of the top 10 embedding models on MLEB, broken down by domain type. Scores represent NDCG@10 averaged across all datasets within each domain (Judicial, Contractual, Regulatory). Legal domain-adapted models (Kanon 2 Embedder, Voyage 3 Large, Voyage 3.5, Voyage Law 2) show particularly strong performance.}
\label{fig:top10}
\end{figure}

We observe that the qualities that make an embedding model perform well at general multilingual information retrieval tasks are not necessarily the same as those that make a model perform well at legal information retrieval. Currently, Gemini Embedding ranks 1st on MTEB and Voyage 3.5 ranks 23rd, whereas on MLEB, Gemini is only 7th and Voyage 3.5 is 3rd.

We also observe that strong performance on MLEB correlates with legal domain adaptation. The top scoring models on MLEB, including Kanon 2 Embedder and Voyage 3 Large and Voyage 3.5, have all been optimized in some way for the legal domain. Kanon 2 Embedder, in particular, was pretrained and finetuned largely on legal documents. Voyage Law 2, despite being an older and likely smaller model than most other high scoring embedding models, ranks 8th, ahead of OpenAI Text Embedding 3 Large. Similar to Kanon 2 Embedder, Voyage Law 2 was optimized specifically for legal information retrieval.

\subsection{Speed}
We assessed the total evaluation time (including network latency) of commercial models on MLEB with a batch size of 16 for documents and 1 for queries (reflective of real-world conditions under which batching of queries is less likely), except when evaluating Voyage AI models on datasets with especially long sequences, in which case a batch size of 1 was used to avoid hitting token rate limits.

\begin{figure}[h!]
\centering
\includegraphics[width=0.9\textwidth]{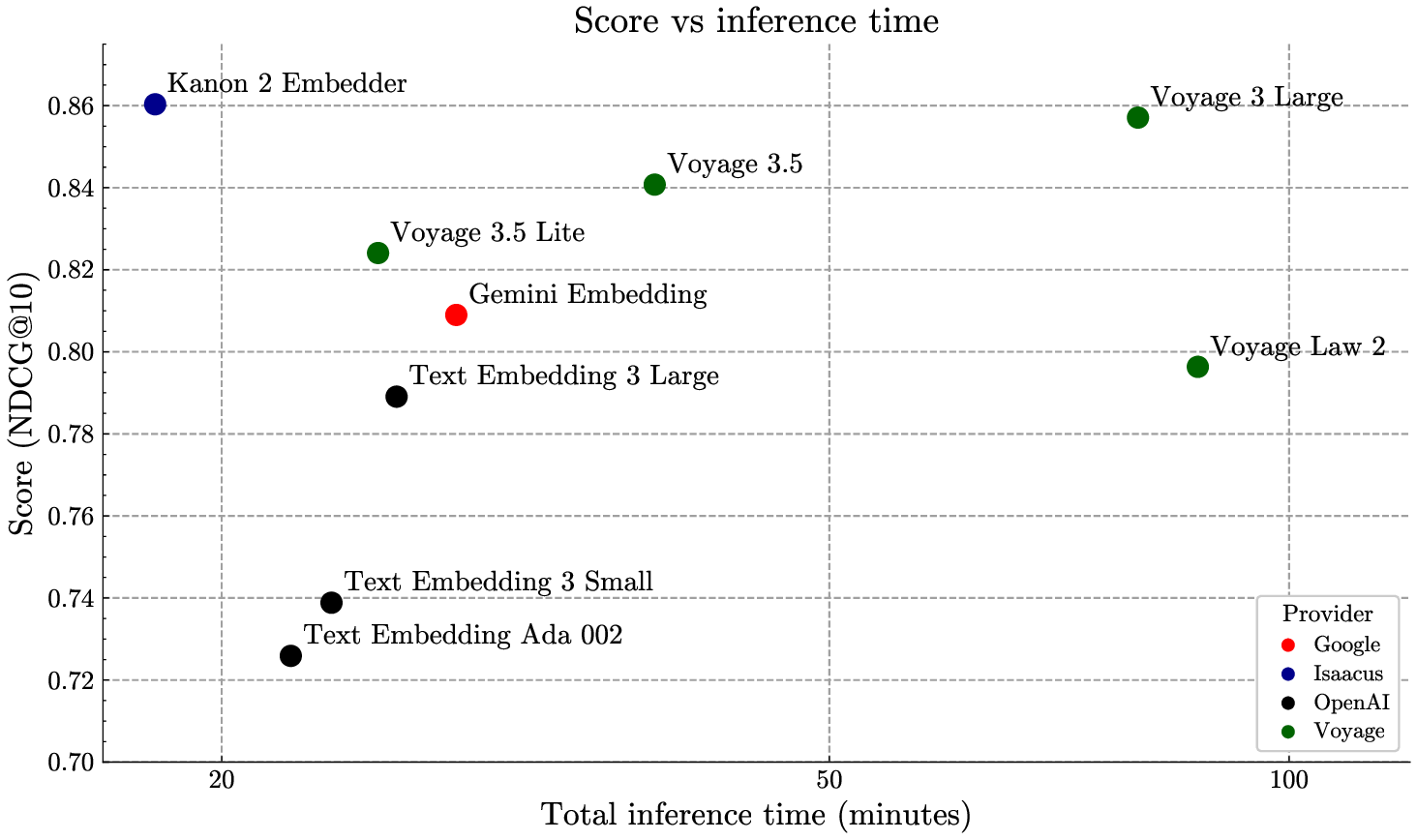}
\caption{NDCG@10 score versus total evaluation time for commercial embedding models on MLEB. Evaluation time includes network latency and reflects real-world conditions (batch size of 16 for documents, 1 for queries). The plot demonstrates the speed-accuracy tradeoff among commercial offerings.}
\label{fig:speed}
\end{figure}

\subsection{Limitations}
Unfortunately, we note that we were unable to evaluate any of Cohere's embedding models as their terms of service expressly forbid benchmarking of any kind. Additionally, we note that there is some potential for data leakage in respect of the models of Voyage AI, Jina, and Google, as their terms of service opt either a portion or all of their API users by default into sharing data for training purposes (which would likely include benchmarks given that customers tend to conduct their own evaluations of models).

\section{Conclusion}
We presented the Massive Legal Embedding Benchmark (MLEB), a high-quality, expert-annotated open-source benchmark for legal information retrieval. Consisting of ten evaluation sets across six jurisdictions (the US, UK, EU, Australia, Ireland, and Singapore) and five document types (cases, legislation, regulatory guidance, contracts, and literature), MLEB is the most comprehensive benchmark for legal embeddings to date. Furthermore, MLEB represents a significant new contribution to the legal information retrieval landscape, with seven datasets being released for the first time in order to increase the representation of jurisdictions and legal domains outside of American law and contract law.

MLEB, its constituent datasets, and evaluation code have all been publicly released under open-source licenses. See the Data Availability section for complete access information. We encourage reproductions of our results and intend to continue improving MLEB with subsequent expansions.

\section{Data Availability}\label{sec:data_availability}
All datasets comprising MLEB are publicly available on Hugging Face (\url{https://huggingface.co/isaacus}) under open-source licenses. The evaluation code and benchmark implementation are available on GitHub (\url{https://github.com/isaacus-dev/mleb}). Direct links to each dataset are provided below:

\begin{itemize}
\item Bar Exam QA\\
\url{https://huggingface.co/datasets/isaacus/mteb-barexam-qa}
\item SCALR\\
\url{https://huggingface.co/datasets/isaacus/mleb-scalr}
\item Singaporean Judicial Keywords\\
\url{https://huggingface.co/datasets/isaacus/singaporean-judicial-keywords}
\item GDPR Holdings Retrieval\\
\url{https://huggingface.co/datasets/isaacus/gdpr-holdings-retrieval}
\item Australian Tax Guidance Retrieval\\
\url{https://huggingface.co/datasets/isaacus/australian-tax-guidance-retrieval}
\item Irish Legislative Summaries\\
\url{https://huggingface.co/datasets/isaacus/irish-legislative-summaries}
\item UK Legislative Long Titles\\
\url{https://huggingface.co/datasets/isaacus/uk-legislative-long-titles}
\item Contractual Clause Retrieval\\
\url{https://huggingface.co/datasets/isaacus/contractual-clause-retrieval}
\item License TL;DR Retrieval\\
\url{https://huggingface.co/datasets/isaacus/license-tldr-retrieval}
\item Consumer Contracts QA\\
\url{https://huggingface.co/datasets/isaacus/mleb-consumer-contracts-qa}
\end{itemize}

\section{Disclosures}
In the interests of transparency, Kanon 2 Embedder was created by Isaacus, a foundational legal AI research company, which also sponsored the creation of MLEB.

\bibliographystyle{plain}
\bibliography{references}

\clearpage

\begin{appendices}

\section{Contractual Clause Types}\label{app:clause_types}

The following table lists all 45 contractual clause types and their corresponding NLI-style statements used in the Contractual Clause Retrieval evaluation set.

\begin{longtable}{>{\raggedright}p{5cm}>{\raggedright\arraybackslash}p{10cm}}
\toprule
\textbf{Type} & \textbf{Statement} \\
\midrule
\endfirsthead

\multicolumn{2}{c}{\tablename\ \thetable\ -- \textit{Continued from previous page}} \\
\toprule
\textbf{Type} & \textbf{Statement} \\
\midrule
\endhead

\midrule
\multicolumn{2}{r}{\textit{Continued on next page}} \\
\endfoot

\bottomrule
\endlastfoot

Set-off clause & This is a contractual provision that permits a contracting party to deduct liabilities owed to it by the counterparty from liabilities it owes to the counterparty. \\
\addlinespace
Authority to sign clause & This is a contractual provision that represents or warrants that a contracting party has authority to bind the entity it represents. \\
\addlinespace
Power of attorney & This is a contractual provision that grants a party a power of attorney to act on behalf of a contracting party. \\
\addlinespace
Termination for cause clause & This is a contractual provision that allows a contracting party to terminate the contract in the event of a breach or default by the counterparty. \\
\addlinespace
Good faith clause & This is a contractual provision that obligates a contracting party to act in good faith. \\
\addlinespace
Payment currency clause & This is a contractual provision that specifies the currency in which payment is to be made. \\
\addlinespace
Termination for change of control clause & This is a contractual provision that permits a contracting party to terminate the contract in the event of a change in the ownership or control of the counterparty. \\
\addlinespace
Conditions precedent clause & This is a contractual provision that specifies conditions to be satisfied for obligations or effectiveness to arise. \\
\addlinespace
Counterparts clause & This is a contractual provision that permits the contract to be executed in separate counterparts. \\
\addlinespace
Workplace surveillance clause & This is a contractual provision that states that an employer may surveil their employee. \\
\addlinespace
No improvements clause & This is a contractual provision that restricts a contracting party from making improvements to property. \\
\addlinespace
Termination for insolvency clause & This is a contractual provision that allows a contracting party to terminate the contract in the event of the bankruptcy or insolvency of the counterparty. \\
\addlinespace
Exclusivity clause & This is a contractual provision that restricts a contracting party from dealing with other parties apart from the counterparty within a particular scope or territory. \\
\addlinespace
Termination for force majeure clause & This is a contractual provision that allows a contracting party to terminate the contract in the event of conditions beyond their control. \\
\addlinespace
Vesting clause & This is a contractual provision that specifies the schedule or conditions under which interests vest. \\
\addlinespace
Termination for convenience clause & This is a contractual provision that allows a contracting party to terminate the contract for any reason. \\
\addlinespace
Drag along clause & This is a contractual provision that permits majority holders to require minority holders to sell their interests on the same terms in a sale. \\
\addlinespace
Cumulative rights clause & This is a contractual provision that specifies rights and remedies are cumulative, not exclusive of those provided by law. \\
\addlinespace
Lock-up clause & This is a contractual provision that restricts a contracting party from selling or transferring securities for a period after issuance or listing. \\
\addlinespace
Break fee clause & This is a contractual provision that requires a contracting party to pay the counterparty in the event of failure to complete a transaction. \\
\addlinespace
Waiver of jury trial clause & This is a contractual provision that waives a contracting party's right to a jury trial. \\
\addlinespace
Return or destruction of materials clause & This is a contractual provision that requires a contracting party to return or destroy material. \\
\addlinespace
Disclaimer & This is a contractual provision that disclaims responsibility for warranties, representations, liabilities or obligations. \\
\addlinespace
Novation clause & This is a contractual provision that permits the substitution of a new party in place of a contracting party. \\
\addlinespace
Confidentiality clause & This is a contractual provision that restricts the use of information protected by a duty of confidence. \\
\addlinespace
Injunctive relief clause & This is a contractual provision that entitles a party to seek injunctive relief. \\
\addlinespace
Moral rights waiver clause & This is a contractual provision that waives a contracting party's moral rights in intellectual property. \\
\addlinespace
Probation clause & This is a contractual provision that permits an employer to terminate an employee without cause or notice during a probationary period. \\
\addlinespace
Escalation to senior management clause & This is a contractual provision that permits the escalation of disputes to senior management. \\
\addlinespace
Limitation of liability for force majeure clause & This is a contractual provision that limits the liability of a contracting party for conditions beyond their control. \\
\addlinespace
Late payment interest clause & This is a contractual provision that requires the payment of interest on overdue liabilities. \\
\addlinespace
Choice of venue clause & This is a contractual provision that specifies the jurisdiction in which disputes over the contract should be resolved. \\
\addlinespace
Clawback clause & This is a contractual provision that entitles a contracting party to recover amounts previously paid to the counterparty. \\
\addlinespace
No assignment clause & This is a contractual provision that prohibits assignment of rights or obligations. \\
\addlinespace
Cure period clause & This is a contractual provision that specifies a period during which a breaching contracting party may remedy or cure their breach. \\
\addlinespace
Penalty clause & This is a contractual provision that imposes a penalty on a contracting party in the event of a failure to perform a contractual obligation. \\
\addlinespace
Suspension clause & This is a contractual provision that entitles a contracting party to suspend their obligations. \\
\addlinespace
Obligatory compliance with all applicable laws clause & This is a contractual provision that obligates a contracting party to comply with all relevant laws. \\
\addlinespace
Further assurances clause & This is a contractual provision that requires a contracting party to take necessary steps or actions to give effect to the contract. \\
\addlinespace
IP assignment clause & This is a contractual provision that assigns intellectual property rights. \\
\addlinespace
Tag along clause & This is a contractual provision that entitles minority holders to participate in a sale by majority holders on the same terms. \\
\addlinespace
Precedence of documents clause & This is a contractual provision that specifies the order of precedence that parts of the contract or other documents take in interpretation of the agreement. \\
\addlinespace
Bonus clause & This is a contractual provision that provides for the payment of bonuses. \\
\addlinespace
Right of first refusal clause & This is a contractual provision that entitles a contracting party to match an offer before an interest is transferred. \\
\addlinespace
Most favored nation clause & This is a contractual provision that requires a contracting party to offer the counterparty goods, services or interests on terms at least as favorable as those granted to anyone else. \\

\end{longtable}

\end{appendices}

\end{document}